\documentclass[aip,cha,sd,amsmath,amsthm,amssymb,reprint]{revtex4-1}
\usepackage{graphicx,epsfig,pstricks,pst-node}
\usepackage{lipsum}
\usepackage{subcaption}
\usepackage{setspace,ulem}
\usepackage{times}
\usepackage{bm}
\usepackage{comment}
\usepackage{multirow}
\usepackage{svg}

\usepackage{kotex}

\begin{document}

\title{Unsupervised Reservoir Computing for Multivariate Denoising of Severely Contaminated Signals}

\author{Jaesung Choi}
{\email{joseph9463@kias.re.kr}}
\affiliation{Center for Artificial Intelligence and Natural Sciences,
	Korea Institute for Advanced Study,
	Seoul 02455, South Korea}

\author{Pilwon Kim*}
{\email{pwkim@unist.ac.kr}}
\affiliation{Department of Mathematical Sciences, 
	Ulsan National Institute of Science and Technology, 
	Ulsan 44919, South Korea}

\date{\today}

\begin{abstract}
The interdependence and high dimensionality of multivariate signals present significant challenges for denoising, as conventional univariate methods often struggle to capture the complex interactions between variables. A successful approach must consider not only the multivariate dependencies of the desired signal but also the multivariate dependencies of the interfering noise. In our previous research [4], we introduced a method using machine learning to extract the maximum portion of ``predictable information" from univariate signal. We extend this approach to multivariate signals, with the key idea being to properly incorporate the interdependencies of the noise back into the interdependent reconstruction of the signal. The method works successfully for various multivariate signals, including chaotic signals and highly oscillating sinusoidal signals which are corrupted by spatially correlated intensive noise. It consistently outperforms other existing multivariate denoising methods across a wide range of scenarios.
\end{abstract}


\maketitle

\section{INTRODUCTION}

Multivariate signal denoising methods aim to enhance data quality by mitigating noise while preserving essential information across multiple dimensions or variables. This is crucial in various domains such as biomedical signal processing, finance, and environmental monitoring, where signals exhibit complex correlations and noise patterns. Denoising techniques span a wide spectrum, from basic adaptive filtering approaches like the least mean square  algorithm [1]  to more advanced multiscale methods leveraging the discrete wavelet transform [2] and empirical mode decomposition [3].

These techniques decompose signals into multiple scales and facilitate the identification and removal of noise components, typically performing by thresholding. However, the signals of interest may contain high frequency components which are often improperly rejected by thresholding the detail coefficients. Chaotic signals are another example that are hard to be captured by thresholding-based techniques, as they typically present across scales in the frequency domain.

A further difficulty especially in multivariate denoising that we want to focus on in this paper is the multivariate dependencies of signals and noise: the signals are commonly inter-coupled and so are the measurement noises. Since their multivariate dependency structures are distinct in general, we better utilize both information to separate them properly. 

The method proposed in this paper extends the single-variable denoising method first introduced in [4] to multivariate signal denoising, by incorporating the aforementioned multivariate dependencies. The approach leverages time series prediction. With the recent progress in machine learning, numerous techniques for time series analysis have emerged, including those utilizing neural networks, enabling the effective identification of deterministic patterns even within individual time series. This redefines the problem of signal filtering as one of signal reconstruction.

By using the given signal as training data, the machine predictor is tasked with uncovering as many predictable patterns as possible within the signal. Provided the predictor is unbiased and its capacity is finely tuned to avoid overfitting, it can accurately extract the deterministic components of the signal. In [4], it was demonstrated that this approach successfully eliminates various types of additive and multiplicative noise, as well as Gaussian and non-Gaussian noise, even when their distributions are not known in advance. 

The essence of multivariate denoising is to find the structural pattern of cross-coupled signals. Training the predictor to reproduce the signals corresponds to this. As the signal and noise are separated in this way, the distribution of the noise is naturally identified. The contribution of our research lies in improving the deterministic reconstruction of the signal by utilizing the multivariate dependency of the noise obtained from this process. In other words, we confirmed that the estimation of the interrelationships of the signal and the estimation of the interrelationships of the noise can both be progressively improved through this iterative process of exchanging results. In various applications, we observed that even a single iteration of this exchange can yield sufficiently good results.

This paper is organized in three main sections. Section 2 proposes the overall algorithm of the method and introduces the echo state network which is used as a predictor in our work.  In Section 3, we explain how the signal can be re-reproduced directly from a tentative estimation of the error distribution. Section 4 illustrates some numerical examples including high-dimensional  and severely contaminated signals and demonstrate the superiority of the proposed method in comparison to other popular multivariate denoising methods.

\section{Signal reconstruction using unsupervised Reservoir computing}
This paper deals with multivariate denoising with regression models of the form
\[
x_i = q_i + \xi_i, \quad 0 \le i \le N,
\]
where the observation \(x_i\) is \(p\)-dimensional, \(q_i\) is the deterministic signal to be recovered and \(\xi_i\) is a spatially correlated noise.

We treat the given signal \(x_i, \; 0 \le i \le N\) to be processed as a dataset for learning, and split it into a training part \(x_i, \; 0 \le i \le K\) and a validation part \(x_i, \; K < i \le N\). We adopt a \(m\)-step time series predictor \(P\) based on a machine learning method which supposedly reproduces one step ahead from previous steps as
\[
x_i = P(x_{i-1}, x_{i-2}, \cdots, x_{i-m}), \quad m \le i \le N
\]
after learning the signal. The denoising procedure consists of the following five steps:
\begin{itemize}
\item\textit{Step 1}: Train the predictor \(P\) to minimize the cost function over the training part,
\[
\sum_{i=m}^{K} \left\| x_i - P(x_{i-1}, x_{i-2}, \cdots, x_{i-m}) \right\|^2.
\]

\item\textit{Step 2}: Reconstruct \(q_i\) and \(\xi_i\) using the predictor \(P\):
\begin{align*}
\hat{q}_i &= P(x_{i-1}, x_{i-2}, \cdots, x_{i-m}),\\
\hat{\xi}_i&=x_i-\hat{q}_i, 
\end{align*}
for $m \le i \le N$.
\item\textit{Step 3}: Apply PCA on the noise covariance matrix \(\mathbb{E}[\hat{\xi}_i \hat{\xi}_i^T] = V \Sigma V^T\).

\item\textit{Step 4}: Compute the \textit{interference calibration matrix} \(\Lambda\). (Section 3).

\item\textit{Step 5}: Reform the Step 1 and 2 again with the substitution \(y = \Lambda V x\) and transform the results back.
\end{itemize}

Steps 1 and 2 correspond to tentative denoising, which extends the univariate signal denoising method applied in [4] to multivariate signals. Steps 3 to 5 pertain to the core methods for improving denoising, and the details are discussed in Section 3.

The predictor \(P\) in the process outlined in the previous section can be implemented with any common machine learning methods. \textit{Reservoir computing} (RC) is one of the popular choices to deal with time series due to its simple architecture and dynamic nature [5]. RC is a suitable predictor for our purpose as it can learn the dynamics from a single time series input.

RC comprises two main components: (i) a \textit{reservoir} is a fixed nonlinear recurrent network and (ii) a \textit{readout} is a trainable linear output layer. We especially use a simple discrete type of RC, Echo State Networks (ESN) to predict the time series \(x_i, i = 0, 1, \cdots\). ESN consists of \(L\) nodes whose temporal states \(r(i) \in \mathbb{R}^L\) evolves through the equation:
\[
r(i+1) = (1 - \alpha)r(i) + \alpha \tanh(Ar(i) + W_{\text{in}} x_{i-1}), \quad i = 1, 2, \cdots
\]
where \(\alpha\) is a leaking rate, \(A \in \mathbb{R}^{L \times L}\) the internal weight matrix, and \(W_{\text{in}} \in \mathbb{R}^{L \times p}\) the input weight matrix. For more general introduction to ESN, the readers are referred to [6, 7].

The readout weight matrix \(W_{\text{out}}\) is determined in the training process by solving:
\[
W_{\text{out}} = \arg \min_{W \in \mathbb{R}^{n_y \times L}} \left( \sum_{i=1}^{K} \left\| x_{i+1} - Wr(i+1) \right\|^2 + \lambda \|W\|_F^2 \right),
\]
where \(\lambda\) is a regularization parameter.

Once \(W_{\text{out}}\) is obtained, we can estimate the deterministic signal as
\[
\hat{q}_i = W_{\text{out}} r(i), \quad i = 0, 1, \cdots, N
\]
to implement the tentative denoising process described in Step 2.

In the subsequent discussion, we call our proposed method as \textit{Multivariate Signal-Separation using Reservoir Computing} (MSSRC). As a final remark on the algorithm of MSSRC, let us remind once more that the predictor $P$ is trained on a single multivariate time series to be filtered, and then attempts to reproduce the very same time series. A large training error can be interpreted as a large measurement noise. Step 1 and 2 can be repeated if necessary, to find the optimal hyperparameters of the predictor $P$ minimizing the training error. In this paper we use  Surrogate optimization, until the training error reaches its minimum.

\section{Interference-calibrated signal recovery}

In Step 3 of the algorithm, we compute the SVD of \(C = \mathbb{E}[\hat{\xi}_i \hat{\xi}_i^T]\) providing an orthogonal matrix \(V\) such that \(C = V \Sigma V^T\) where \(\Sigma = \text{diag}(\sigma_1, \sigma_2, \cdots, \sigma_p)\) and \(V = [v_1|v_2| \cdots |v_p]\). Here the singular value \(\sigma_k\) corresponds to variance of the noise in the direction of \(v_k\), the \(k\)-th principal component of the signal.

What we aim to do in Steps 4 and 5 is to re-perform the reproduction of signals, which was tentatively attempted in Steps 1 and 2, using new bases \(v_1, v_2, \cdots, v_p\). Namely, we try to predict \(Vx_i\) from \(Vx_{i-1}, Vx_{i-2}, \cdots, Vx_{i-m}\) by training the predictor \(P\). Of course, since \(V\) is an orthogonal transformation, this transformation itself brings little additional benefit to the prediction. To make a substantial difference in the second reproduction, we note that the \(p\) directions in this input, namely \(Vx = (v_1^T x)v_1 + (v_2^T x)v_2 + \cdots + (v_p^T x)v_p\), contribute differently to the reconstruction process. In other words, the strength of the noise in each direction determines how much the information of the signal \(x\) in that direction contributes to the reconstruction when applied to the predictor. To investigate this, we define the directional variance of the reproduced signal as
\[
\sigma_k^S = \text{Var}[v_k^T \hat{q}_i]
\]
where \(\hat{q}_i\) is a signal tentatively reproduced at Step 2.

The ratio of \(\sigma_k^S\) to \(\sigma_k\) roughly indicates directional SNR. This ratio being low implies the deterministic portion of the signal in that direction is small, and likely contributes less to reconstruction of the whole signal. For example, if \(\sigma_k\) is excessively larger than \(\sigma_k^S\), we can attain almost no information on \(x_i\) from \(v_k^T x_{i-1}\). To describe this tendency, we use the reconstruction contribution weight in the direction of \(v_k\) as
\[
w_k = \frac{1}{1 + \frac{\sigma_k}{\sigma_k^S}}.
\]
Note \(w_k \to 1\) as \(\sigma_k \to 0\) and \(w_k \to 0\) as \(\sigma_k \to \infty\).

Now, the \textit{interference calibration matrix} is defined as \(\Lambda = \text{diag}(w_1, w_2, \cdots, w_n)\) for scaling according to noise intensity. We then substitute \(y_i = \Lambda V x_i\) and reapply the reproducing process on signal \(y_i\). The interference calibration matrix is used to adjust the influence of each component in the signal reconstruction process, mitigating the varying degrees of interference caused by different noise components. In the next section, it is shown that this second reproduction under the substitution leads to outstanding improvement. 

\section{Numerical experiments}

\begin{figure*}[t]
    \centering
    \includegraphics[width=\textwidth]{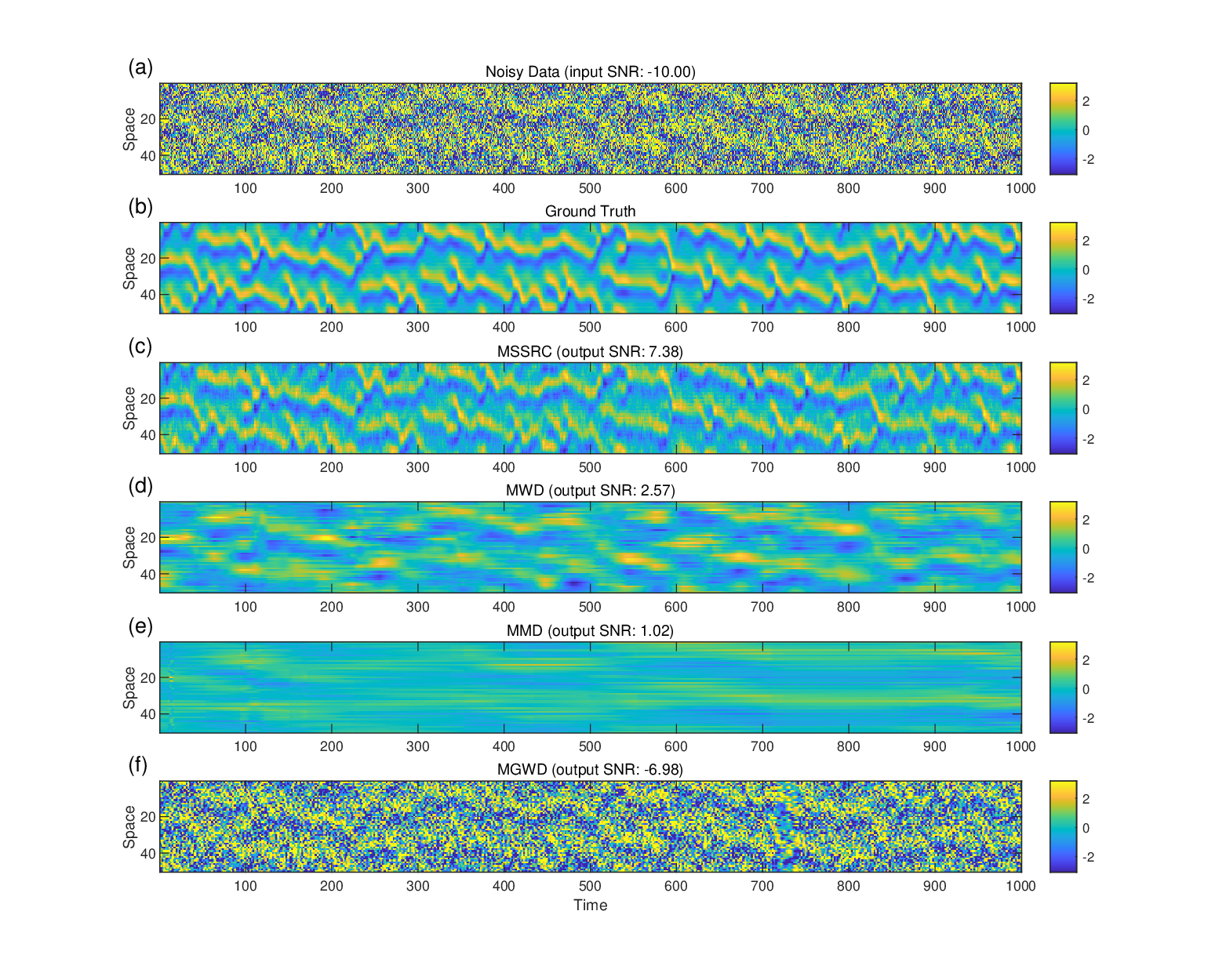}
    \caption{Time plots of (a) noisy version of KS and (b) ground truth along with the its denoised versions obtained from (c) the proposed MSSRC, (d) MWD, (e) MMD and (f) MWGD }
    \label{fig:kssignals}
\end{figure*}

\begin{figure*}[t]
    \centering
    \includegraphics[width=\textwidth]{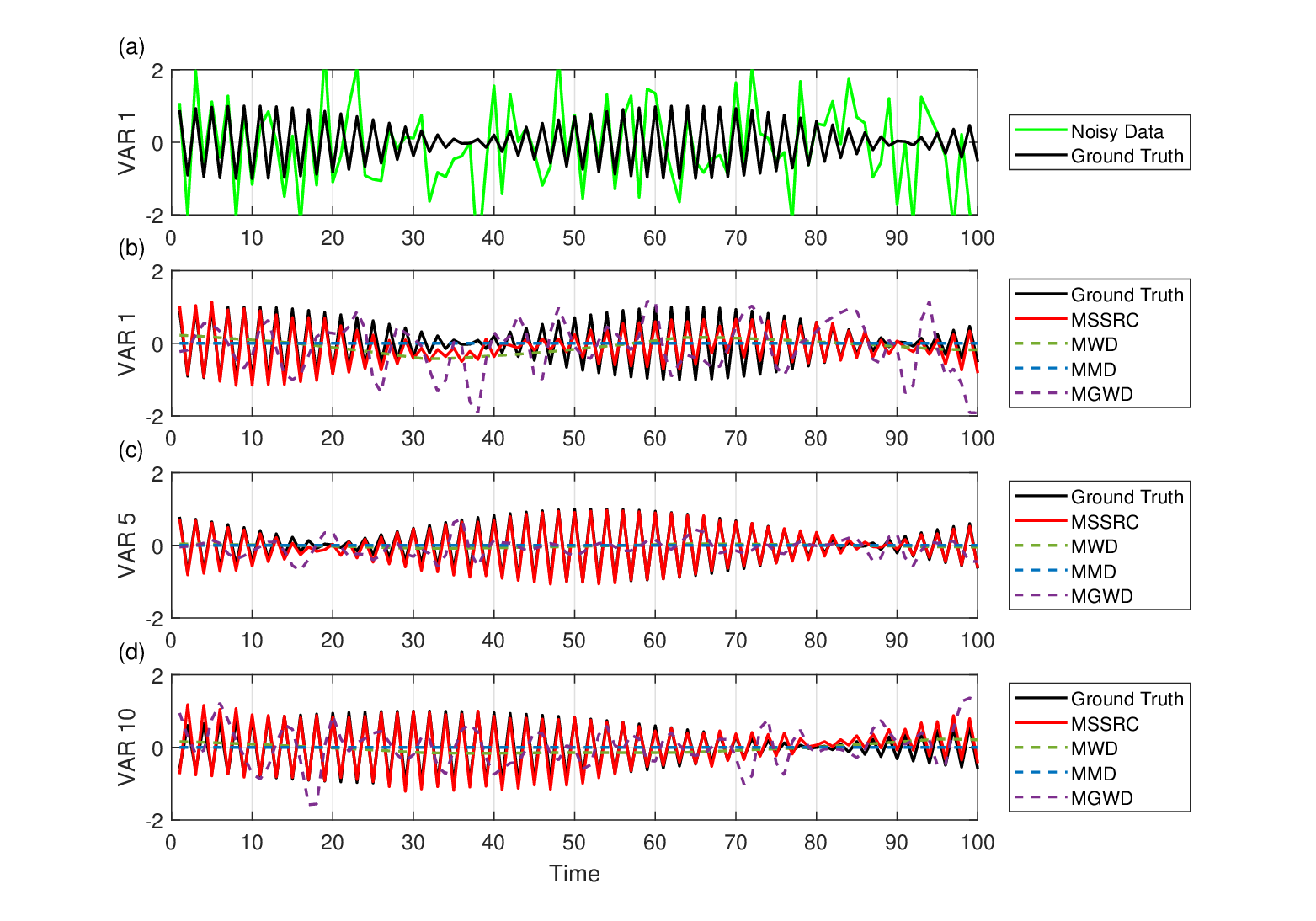}
    \caption{Time plots of high-frequency sinusoidal signals before and after denoising. (a) a variable before denoising and (b),(c) and (d) variables after denoising. }
    \label{fig:hfsignals}
\end{figure*}

This section applies the separation procedure described in Section 2 and 3 to various type of signals. We assess the effectiveness of the proposed method by comparing its performance with the current leading techniques in multivariate signal denoising: MWD [8], MMD [9,13], and MGWD [10,14]. We are especially interested in chaotic and highly oscillating target signal , which is often hard to identify from noise. More specifically, we consider two types of deterministic systems: a 50-dimensional system based on the Kuramoto-Sivashinsky equation (KS) and a 10-dimensional high-frequency sinusoidal signal. For the KS system, we use numerical solutions of the PDE sampled at multiple spatial points. Refer to Supplement for more details of the signals including the generating equations. We evaluate the denoising performance by measuring the signal-to-noise ratio (SNR) of the denoised signal. SNR values are reported for each channel individually, along with the average output SNR across all channels.

\subsection{Kuramoto-Sivashinsky} 
We begin by examining artificially generated time series produced through numerical simulation of the Kuramoto-Sivashinsky equation, a partial differential equation system known for revealing spatiotemporal chaos. Figure 1 shows the denoised data obtained after applying the proposed and comparative methods to original noisy KS signal severely contaminated with input SNR=-10.  Note that, unlike MWD (output SNR=2.57), MMD (1.02) and MWGD (-6.98), the denoised signal obtained from MSSRC closely follows the ground truth for all input channels with the output SNR 7.38.

\begin{table*}[t]
    \centering
    \includegraphics[width=\textwidth]{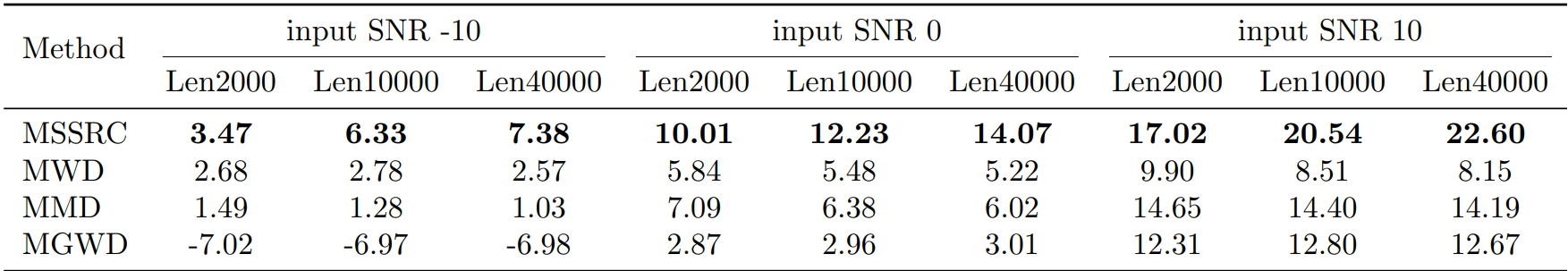}
    \caption{Average output SNR from various comparative multivariate signal denoising methods on Kuramoto-Sivashinsky signals with various setting}
    \label{tab:table1}
\end{table*}

\begin{table*}[t]
    \centering
    \includegraphics[width=\textwidth]{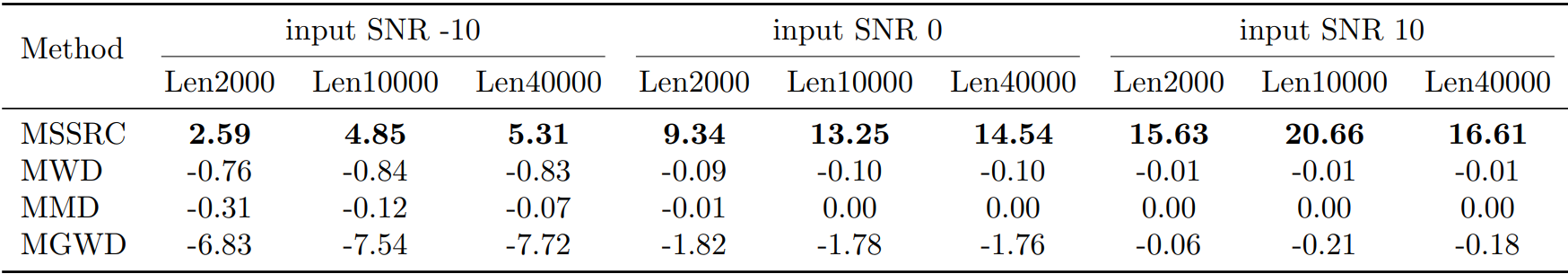}
    \caption{Average output SNR from various comparative multivariate signal denoising methods on high-frequncy multivariate sinusoidal signals with various setting}
    \label{tab:table2}
\end{table*}

Table I reports the values of average output SNRs of denoised KS signals obtained from MWD, MMD, MGWD and the proposed MSSRC method for a range of input SNRs (−10 dB to 10 dB) and time series length (2000 to 40000). In all cases, the proposed method MSSRC outperforms the others, with its superiority becoming more pronounced as the noise level increases. Additionally, while the denoising performance of the other methods tends to be independent of the length of the given time series, MSSRC, which treats the time series to be filtered as training data, exhibits improved performance as the time series lengthens. That is, even at the same noise level, the performance gap widens as the time series becomes longer.

\subsection{High-frequency sinusoidal signal}
Most conventional denoising methods have difficulties in identifying high-frequency deterministic information in the signal from random noise [4]. For the second experiments, we use the 10 dimensional sinusoidal signals with various high frequency under spatially correlated noise. In Figure 2, comparative methods are applied on the noisy signals corresponding to input SNR=0 and the resulting denoised signals are displayed. The one of the original noisy signals is plotted in (a) to show difficulties in identifying the high-frequency sinusoidal signal hidden in intense noise. As we can see in Figure 2(b), (c) and (d), the proposed method MSSRC successfully reproduces the ground truth signals, while the other comparative methods show poor performances. 

Table 2 summarizes the values of average output SNRs of denoised high-frequency sinusoidal signals obtained from MWD, MMD, MGWD and the proposed MSSRC method for a range of input SNRs (−10 dB to 10 dB) and time series length (2000 to 40000). Once again, the proposed method MSSRC outperforms the others in all cases. As in the previous KS case, MSSRC was excellent at recovering target signals with very low SNR. We also confirmed that filtering performance tended to improve as the length of the time series increased. 

\section{Discussion}

The proposed method is flexible and simple, and at its core it exploits the ability of machine learning to maximize the extraction of deterministic patterns. The effect of noise is neutralized or eliminated in the process of reproducing deterministic patterns in the signal. In extension to multivariate signals, we applied principal component analysis (PCA) to the estimator of the noise distribution and incorporate the interdependencies of the noise back into the interdependent reconstruction of the signal.  

There have been other approaches using PCA in denoising multivariate signals [8, 11, 12]. While those methods apply PCA to the original or filtered signals to enhance the deterministic relationships between the signals, we rather apply it to the tentative noise distribution to find the major axis of variance. By identifying the noise intensity in each direction through PCA, it is possible to assign different scaling factors accordingly on each direction. This enables retraining the predictor to reproduce the signal, resulting in significantly improved filtering outcomes.

In this work, we applied the proposed method to multivariate signals contaminated by Gaussian noise. It was demonstrated in [4] that the same denoising approach using reservoir computing works greatly for a wide range of univariate signals under additive/multiplicative and Gaussian/Non-Gaussian noise. Although not explicitly addressed in this paper, preliminary observations suggest our method remains effective in various cases of multivariate signals with non-Gaussian noise. However, we anticipate further performance improvements for multivariate signals with general noise distributions are achievable by developing more sophisticated manipulation methods beyond PCA, which can better incorporate general noise characteristics into signal enhancement.

\bigskip

\noindent{\bf Data availability}\\
The authors declare that the data supporting the findings of this study can be recreated as described in the manuscript and also obtainable from the corresponding author upon request.\\

\noindent{\bf Acknowledgements}\\
This work is supported by the Center for Advanced Computation at Korea Institute for Advanced Study


\noindent{\bf Competing interests}\\
Authors declare no competing interests.\\

\noindent {\bf Correspondence} and requests for materials should be addressed to Pilwon Kim.\\

\end{document}